\documentclass{bmvc2k}

\usepackage{array}

\title{Weakly-supervised Salient Instance Detection}

\addauthor{Xin Tian}{xtian@mail.dlut.edu.cn}{12}
\addauthor{Ke Xu}{kkangwing@mail.dlut.edu.cn}{12}
\addauthor{Xin Yang}{xinyang@dlut.edu.cn}{1}
\addauthor{Baocai Yin}{ybc@dlut.edu.cn}{13}
\addauthor{Rynson W.H. Lau}{rynson.lau@cityu.edu.hk}{2}

\addinstitution{
 Computer Science Department\\
 Dalian University of Technology\\
 Dalian, China
}
\addinstitution{
 Computer Science Department\\
 City University of Hongkong\\
 Hongkong SAR, China
}
\addinstitution{
 Pengcheng Lab\\
 Shenzhen, China
}

\runninghead{Xin \etal}{Weakly-supervised Salient Instance Detection}

\newcommand{\kk}[1]{\textcolor[rgb]{0.0,0.0,0.0}{#1}}
\newcommand{\tx}[1]{\textcolor[rgb]{0,0,0}{#1}}

\def\eg{\emph{e.g}\bmvaOneDot}

\def\etal{\emph{et al}\bmvaOneDot}
\def\ie{\emph{i.e}\bmvaOneDot}

\usepackage{verbatim}

\begin{document}

\maketitle
\begin{abstract}
Existing salient instance detection (SID) methods typically learn from pixel-level annotated datasets. In this paper, we present the first weakly-supervised approach to the SID problem. Although weak supervision has been considered in general saliency detection, it is mainly based on using class labels for object localization. However, it is non-trivial to use only class labels to learn instance-aware saliency information, as salient instances with high semantic affinities may not be easily separated by the labels.
We note that subitizing information provides an instant judgement on the number of salient items, which naturally relates to detecting salient instances and may help separate instances of the same class while grouping different parts of the same instance.
Inspired by this insight, we propose to use class and subitizing labels as weak supervision for the SID problem. We propose a novel weakly-supervised network with three branches:
a Saliency Detection Branch leveraging class consistency information to locate candidate objects;
a Boundary Detection Branch exploiting class discrepancy information to delineate object boundaries;
and a Centroid Detection Branch using subitizing information to detect salient instance centroids.
This complementary information is further fused to produce salient instance maps.
%
%
We conduct extensive experiments to demonstrate that the proposed method plays favorably against carefully designed baseline methods adapted from related tasks.
\end{abstract}

\section{Introduction}
\label{sec:intro}
Salient Object Detection (SOD) is a long-standing vision task that aims to segment visually salient objects in a scene. It often serves as a core step for downstream vision tasks like video object segmentation~\cite{wang2015saliency}, object proposal generation~\cite{alexe2012measuring}, and image cropping~\cite{wang2017deep}. Recent deep learning-based SOD methods have achieved a significant performance progress~\cite{wang2017learning,zhuge2018boundary,xu2019structured,su2019selectivity,zhao2019pyramid,hou2017deeply,wang2018detect}, benefited from the powerful representation learning capability of neural networks and large-scale pixel-level annotated training data.
Since annotating pixel-level labels is extremely tedious, there are some works~\cite{wang2017learning,zeng2019multi} that aim to explore cheaper image-level labels (\eg, class labels) to train SOD models in a weakly-supervised manner.
%

Salient Instance Detection (SID) goes further from SOD as it is to identify each salient instance. \kk{This instance-level saliency information can further benefit vision tasks that requires fine-grained scene understanding, \eg, image captioning~\cite{karpathy2015deep}, image editing~\cite{chen2009sketch2photo} and semantic segmentation~\cite{fan2018associating}. However, existing SID methods ~\cite{fan2019s4net,li2017instance,zhang2016unconstrained} still rely on large-scale annotated ground truth masks in order to learn how to segment salient instances with their boundaries delineated.}
Hence, it is worthwhile to study the SID problem from the weakly-supervised perceptive of using cheaper image-level labels.
%

A straightforward solution may be to use class labels to train a weakly-supervised SID model. However, using just class labels to learn a SID model is non-trivial for two reasons. First, class labels can help detect semantically predominant regions~\cite{zhou2016learning}, but these regions are not guaranteed to be visually salient. Second, objects of the same class may not be easy distinguished due to their high semantic affinities.
%
%
We observe that subitizing is naturally related to saliency instance detection. By predicting the number of salient objects, it can serve as global supervision that can help separate instances of the same class and cluster parts of an instance with diverse appearances into one.

Inspired by the above insight, we propose to learn a Weakly-supervised SID network (denoted WSID-Net) using class and subitizing labels.
Our WSID-Net consists of three synergic branches:
%
a salient object detection branch and a boundary detection branch are proposed to locate candidate salient objects and delineate their boundaries, by exploiting semantics from the class labels;
a centroid detection branch is proposed to detect the centroid of each salient instance, by leveraging saliency cues from the subitizing labels. This information is fused to obtain the salient instance maps.
%
%
To demonstrate the effectiveness of the proposed model, we compare it with a variety of baselines adapted from related tasks on the standard benchmark~\cite{li2017instance}.

To summarize, this paper has three main contributions: 1) To the best of our knowledge, we propose the first weakly-supervised method for salient instance detection, which only requires image-level class and subitizing labels \tx{to obtain salient instance maps}; 2) We propose a novel \tx{network (}WSID-Net\tx{)}, with a novel centroid-based subitizing loss to exploit salient instance number information, and a novel Boundary Enhancement module to learn  instance boundaries; 3) We conduct extensive experiments to analyze the proposed method, and verify its superiority against baselines adapted from related state-of-the-art approaches.
\vspace{-5mm}

\section{Related Work}

{\bf Salient Instance Detection (SID).} Existing SID methods are fully-supervised.
Zhang~\etal~\cite{zhang2016unconstrained} propose to detect salient instances with bounding boxes, and propose a {MAP}-based optimization framework to regress a large amount of pre-defined bounding boxes into a compact number of instance-level bounding boxes of high confidences.
However, their method based on bounding boxes cannot detect salient instances with accurately delineated boundaries.
Other works predict pixel-wise masks for the detected salient instances, and typically rely on large amount of manually annotated ground truth labels.
Specifically, Li~\etal~\cite{li2017instance} propose to first predict the saliency mask and instance-aware saliency contour, and then use the existing Multi-scale Combinatorial Grouping (MCG) algorithm~\cite{APBMM2014} to extract instance-level masks.
Fan~\etal~\cite{fan2019s4net} propose an end-to-end SID network based on the object detection model FPN~\cite{lin2017feature}, with a segmentation branch to segment the salient instances.

Unlike these existing SID methods, we propose in this paper to train a weakly-supervised network, which only requires two image-level labels, \ie, the class and subitizing labels.
\vspace{0.1in}

\noindent {\bf Salient Object Detection (SOD).} SOD methods aim at generally detecting salient objects in a scene without differentiating the detected instances.
Liu~\etal~\cite{liu2010learning} formulate the SOD task as a binary segmentation problem for segmenting out the visually conspicuous objects of an image via color and contrast histogram based priors.
Traditional methods propose to leverage different hand-crafted priors to detect salient objects, \eg, image colors and luminance~\cite{achanta2009frequency}, global and local contrast priors~\cite{perazzi2012saliency,cheng2014global}, and {background geometric distance prior}~\cite{yang2013saliency}.
Recently, deep learning based SOD methods achieve superior performance on standard SOD {benchmarks}~\cite{yang2013saliency,li2014secrets,shi2015hierarchical,jiang2013salient,wang2017learning,cheng2014global}, by incorporating salient boundary knowledge \cite{zhuge2018boundary,xu2019structured,su2019selectivity}, fusing deep features~\cite{zhao2019pyramid,hou2017deeply}, and designing attention mechanisms~\cite{wang2018detect,zhang2018bi}.
Particularly, He \etal \cite{he2017delving} propose to leverage numerical representation of subitizing to enrich spatial representations of salient objects.
These methods are typically benefitted from the powerful learning ability of deep neural networks as well as large-scale annotated ground truth data.
To alleviate the data annotation efforts, some methods~\cite{wang2017learning,zeng2019multi} propose to train weakly-supervised deep models using object class labels and class activation maps (CAMs)~\cite{zhou2016learning}.
On the other hand, Li \etal~\cite{li2018contour} propose to leverage pre-trained contour network to generate pseudo labels for training the saliency detection network.

However, existing weakly-supervised SOD methods cannot be directly applied to our problem, as class labels cannot provide instance-level information. In this paper, we propose to use class and subitizing labels to train our SID model.

\vspace{0.1in}
\noindent \tx{
{\bf Weakly-supervised Semantic Instance Segmentation (SIS).} SIS methods aim to detect all instances in a class-specific manner. Although they do not consider the saliency attribute of instances, they are related to our task as they try to segment the objects in an image into instances. Here, we briefly summarize latest weakly-supervised SIS methods, which are adopted as baseline methods for our task.
Based on pixel affinities extracted from the class activation map, IRN \cite{jiwoon2019weakly} learns to predict object seeds and boundaries that can be used to infer the entire region of the target instance.
PRM \cite{zhou2018weakly} first learns to predict peak response maps within class responses, where each peak is generally related to an instance. It then adopts off-the-shelf segment proposals \cite{APBMM2014} to obtain each instance based on the peaks.
In comparison to PRM, PRM+D \cite{cholakkal2019object} further incorporates per-class object number information to learn better spatial distribution of peak-represented instances.
Some other methods \cite{zhu2019learning,laradji2019masks} propose to refine the results of PRM~\cite{zhou2018weakly} in an online way, via jointly learning from class labels and off-the-shelf segment proposals.
}

\vspace{-4mm}

\section{Methodology}

\kk{Class labels are widely explored in weakly-supervised SOD methods for learning to localize candidate objects, based on the pixel-level semantic affinities derived from the network responses to the class labels. However, class labels lack instance-level information, causing over- and under-detection when salient instances are from the same category.
We note that subitizing, a cheap image-level label that denotes the number of salient instances of a scene, can serve as a complementary supervision to the class labels.
Hence, we propose to use both class and subitizing labels to address our weakly-supervised SID problem.
}

\begin{figure}[h]
\begin{center}
\includegraphics[width=0.95\linewidth]{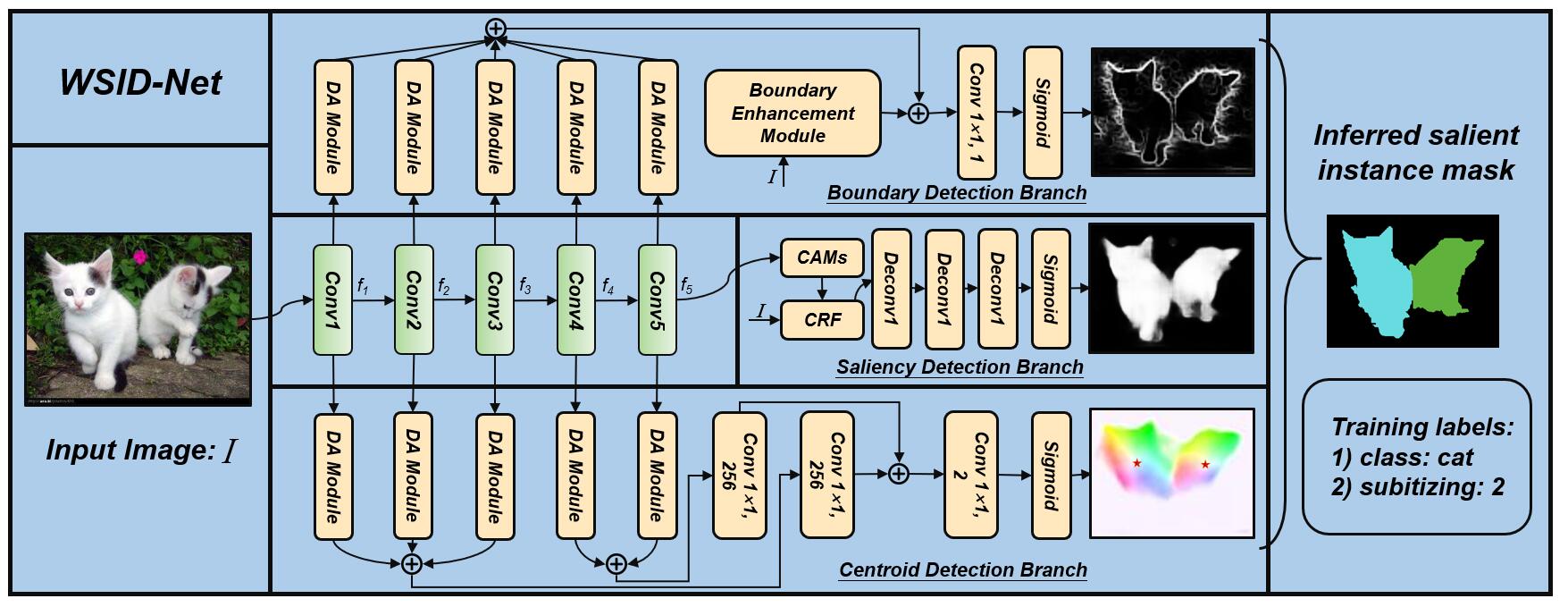}
\end{center}\vspace{-5mm}
\caption{Pipeline overview. Our SID model is trained only using image-level class and subitizing labels. It has three synergic branches: (1) a Boundary Detection Branch for detecting object boundaries using class discrepancy information; (2) a Saliency Detection Branch for detecting objects using class consistency information; (3) a Centroid Detection Branch for detecting salient instance centroids using subitizing information. A random walk method is further applied to fuse these information to obtain final salient instance mask.
}
\label{fig:pipeline}
\vspace{-5mm}
\end{figure}

To this end, we propose a Weakly-supervised SID network (WSID-Net), as shown in Figure~\ref{fig:pipeline}. WSID-Net has three branches: a {$Salieny$ $Detection$ $Branch$} for locating candidate salient objects;
a {$Centroid$ $Detection$ $Branch$} for detecting the centroids of salient instances, where subitizing knowledge is utilized in a novel loss function to provide regularization on the global number of instance centroids;
and a {$Boundary$ $Detection$ $Branch$} for delineating salient instance boundaries, where a novel Boundary Enhancement (BE) module is introduced to resolve the discontinuity problem of detected boundaries.
A novel Double Attention (DA) module is further incorporated to learn the context information for detecting centroids and boundaries.

\subsection{Centroid Detection Branch}\label{sec:322}

\kk{Detecting object centroids is crucial to separating the objects in a weakly-supervised scheme. Unlike existing semantic (instance) segmentation methods~\cite{jiwoon2019weakly,Neven2019InstanceSB,zhou2018weakly,cholakkal2019object,zhu2019learning,laradji2019masks} that detect the centroids based on network responses to the class labels, we propose to introduce subitizing information to explicitly supervise the salient centroid detection process.
}

{\bf Network structure.} We adopt the image-to-image translation scheme, where our network outputs a 2D centroid map, of which the values of each pixel location indicate the offset vector to its instance centroid. The bottom part of Figure~\ref{fig:pipeline} shows the network structure of our centroid detection branch. Given an input image, we first extract multi-scale backbone features $f_{1}$ to $f_{5}$ and feed them to the DA modules for refinement \tx{(to be discussed in Section~\ref{sec:da})}. \tx{The refined features are denoted as ${f_{1}}^{'}$ to ${f_{5}}^{'}$.} We then fuse the high-level features to obtain $f_{h}$: $f_{h} = Conv(Concat(\tx{{f_{3}}^{'},{f_{4}}^{'},{f_{5}}^{'}}))$, which is further fused with the low-level features to produce the centroid map $\mathcal{V}$: $\mathcal{V} = \sigma(Conv(Conv(Concat(f_{h},\tx{{f_{1}}^{'},{f_{2}}^{'}}))))$.

{\bf Centroid-based Subitizing loss.} \kk{It has been shown that penalizing the centroid loss~\cite{jiwoon2019weakly,Neven2019InstanceSB} helps cluster local pixels with high semantic affinities. However, it typically fails when salient instances from the same object category have varying shapes and appearances.}
The reason is that the clustering process of local pixels lacks global saliency supervision on it. Hence, we introduce the centroid-based subitizing loss $\mathcal{L}_{\mathcal{SU}}$ to resolve this problem. We use subitizing to explicitly supervise the number of predicted centroids, which helps constrain the pixel clustering process. We use the Mean Square Error (MSE) to measure $\mathcal{L}_{\mathcal{SU}}$:
\begin{equation}
\mathcal{L}_{\mathcal{SU}} = MSE(t_{\prod_{\mathcal{V}(x_{i})}{x_{i} \in \mathcal{S}}}, t^{*}),
\label{equa:5}
\end{equation}
where $t^{*}$ is the subitizing information, $\mathcal{S}$ denotes the predicted saliency region. ${{\prod_{\mathcal{V}(x_{i})}{x_{i} \in \mathcal{S}}}}$ denotes the predicted offset vectors in the saliency region. $t_{\prod_{\mathcal{V}(x_{i})}{x_{i} \in \mathcal{S}}}$ denotes the \tx{number of} predicted centroid extracted from the offset vectors of the pixels in the saliency region. The loss $\mathcal{L}_{\mathcal{SU}}$ only backpropagates to update the offset vectors in the saliency region,  avoiding the learning process of instance centroid detection  being distracted by the non-salient background.

\begin{figure}[!h]
\centering
\begin{minipage}[t]{0.13\textwidth}
\centering
\scriptsize{\textbf{input\\image}}
\end{minipage}
\begin{minipage}[t]{0.13\textwidth}
\centering
\scriptsize{\textbf{ground \\truth}}
\end{minipage}
\begin{minipage}[t]{0.13\textwidth}
\centering
\scriptsize{\textbf{centroid\\(w/o $\mathcal{L}_{\mathcal{SU}}$)}}
\end{minipage}
\begin{minipage}[t]{0.13\textwidth}
\centering
\scriptsize{\textbf{instance mask\\(w/o $\mathcal{L}_{\mathcal{SU}}$)}}
\end{minipage}
\begin{minipage}[t]{0.13\textwidth}
\centering
\scriptsize{\textbf{centroid\\(w/ $\mathcal{L}_{\mathcal{SU}}$)}}
\end{minipage}
\begin{minipage}[t]{0.13\textwidth}
\centering
\scriptsize{\textbf{instance mask\\(w/ $\mathcal{L}_{\mathcal{SU}}$)}}
\end{minipage}
\centering
\\
\centering
\includegraphics[width=.13\textwidth]{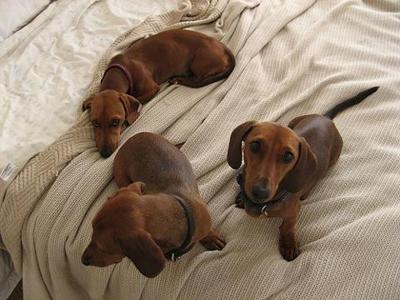}
\includegraphics[width=.13\textwidth]{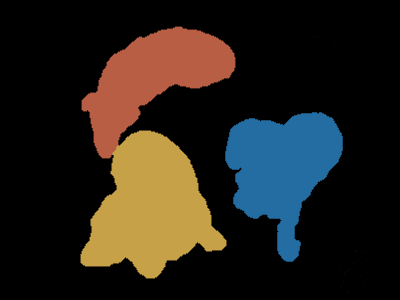}
\includegraphics[width=.13\textwidth]{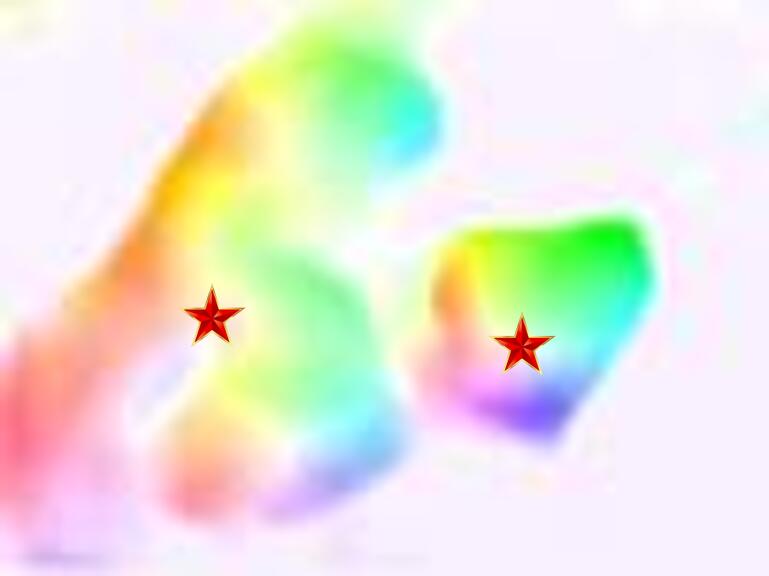}
\includegraphics[width=.13\textwidth]{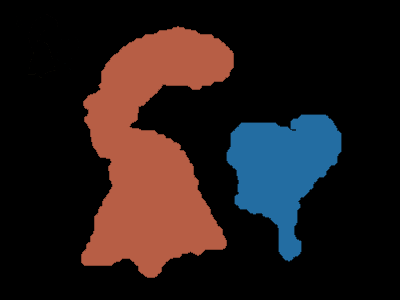}
\centering
\includegraphics[width=.13\textwidth]{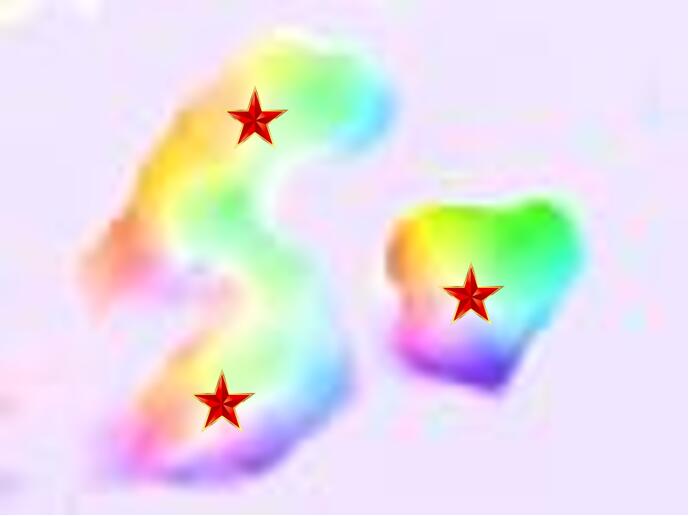}
\includegraphics[width=.13\textwidth]{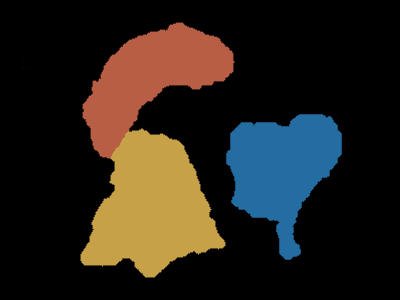}
\caption{Visualization of the centroid detection branch with and without $\mathcal{L}_{\mathcal{SU}}$.}
\label{fig:wsu}
\vspace{-7mm}
\end{figure}

Figure~\ref{fig:wsu} visualizes the results from centroid detection and the corresponding instance segmentation, with and without using the centroid-based subitizing $\mathcal{L}_{\mathcal{SU}}$ loss function. We can see that the network groups the two dogs into one when not using $\mathcal{L}_{\mathcal{SU}}$, as these two dogs have similar appearances and lie next to each other (columns 3 and 4). By introducing $\mathcal{L}_{\mathcal{SU}}$, the network is able to predict a correct number of centroids, and generate reasonable salient instance masks compared with the ground truth (columns 5 and 6).

\subsection{Boundary Detection Branch}\label{sec:321}
\kk{Boundaries provide strong cues for separating salient instances. Unlike fully-supervised SID methods that learn boundary-aware information based on pixel-level ground truth masks, we propose the Boundary Enhancement module to leverage the Canny prior~\cite{john1986} to delineate continuous instance boundaries.}
%

{\bf Network structure.} The top part of Figure~\ref{fig:pipeline} shows the architecture of the boundary detection branch. Given an input image $\mathcal{I}$, \tx{we obtain refined backbone features (${f_{1}}^{'}$ to ${f_{5}}^{'}$) using DA modules (to be discussed in Section~\ref{sec:da})} before they are concatenated and computed to predict the boundary map. {We also feed the input image into the BE module to obtain enhanced edge features $f_{b}$.} The output boundary map $\mathcal{B}$ is then computed as:
$\mathcal{B} = \sigma(Conv(Concat(\tx{{f_{1}}^{'},...,{f_{5}}^{'},f_{b}})))$, where $\sigma$ is the sigmoid activation function.

\begin{figure}[h]
   \begin{center}
   \includegraphics[width=0.85\linewidth]{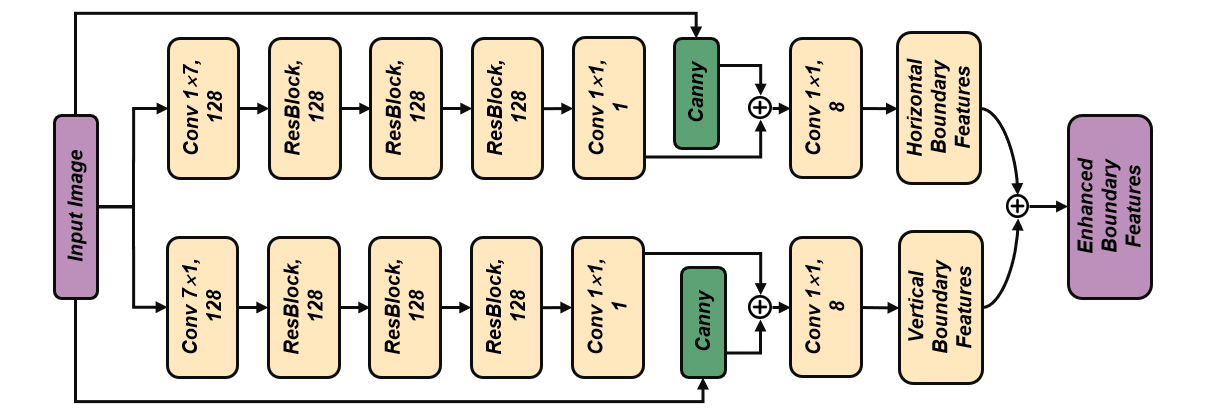}
   \end{center}
   \vspace{-4mm}
   \caption{Illustration of our novel Boundary Enhancement module.}
   \label{fig:BE}
   \vspace{-1mm}
\end{figure}

{\bf BE module.} We apply a random walk algorithm to search a salient instance from a centroid to its boundary. However, it may fail when part of the boundary is discontinuous as the random walk algorithm will also search the region outside the boundary. Hence, we propose the BE module to incorporate the edge prior for learning continuous instance boundaries, as shown in Figure~\ref{fig:BE}. Specifically, we first extract low-level features along the horizontal and vertical directions from the input image, by two $1\times7$ and $7\times1$ convolution layers. These low-level features are then fed into three Residual Blocks~\cite{he2016deep} for feature refinement, which are further concatenated with enriched edges computed from the Canny operator~\cite{john1986}. To compute the final enriched boundary features, another 1$\times$1 convolution layer is applied.

\begin{figure}[h]
\centering
\begin{minipage}[t]{0.09\textwidth}
\centering
\scriptsize{\textbf{input\\image}}
\end{minipage}
\begin{minipage}[t]{0.09\textwidth}
\centering
\scriptsize{\textbf{boundary\\(w/o BE)}}
\end{minipage}
\begin{minipage}[t]{0.09\textwidth}
\centering
\scriptsize{\textbf{instance mask\\(w/o BE)}}
\end{minipage}
\begin{minipage}[t]{0.09\textwidth}
\centering
\scriptsize{\textbf{boundary\\(w/ BE)}}
\end{minipage}
\begin{minipage}[t]{0.09\textwidth}
\centering
\scriptsize{\textbf{instance mask\\(w/ BE)}}
\end{minipage}
\centering
\begin{minipage}[t]{0.09\textwidth}
\centering
\scriptsize{\textbf{input\\image}}
\end{minipage}
\begin{minipage}[t]{0.09\textwidth}
\centering
\scriptsize{\textbf{boundary\\(w/o BE)}}
\end{minipage}
\begin{minipage}[t]{0.09\textwidth}
\centering
\scriptsize{\textbf{instance mask\\(w/o BE)}}
\end{minipage}
\begin{minipage}[t]{0.09\textwidth}
\centering
\scriptsize{\textbf{boundary\\(w/ BE)}}
\end{minipage}
\begin{minipage}[t]{0.09\textwidth}
\centering
\scriptsize{\textbf{instance mask\\(w/ BE)}}
\end{minipage}
\\
\centering
\includegraphics[width=.09\textwidth]{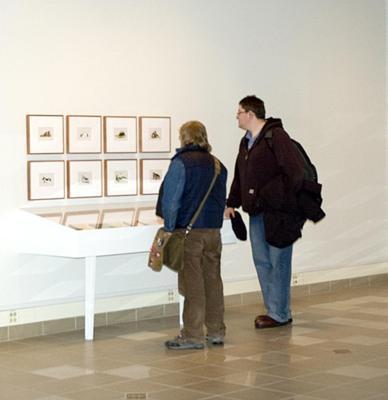}
\includegraphics[width=.09\textwidth]{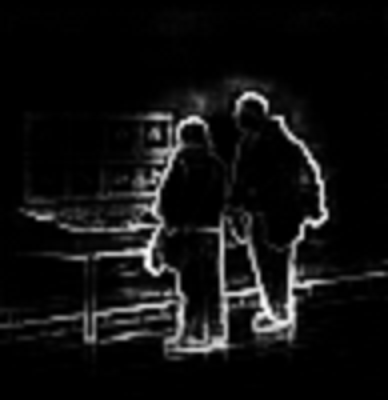}
\includegraphics[width=.09\textwidth]{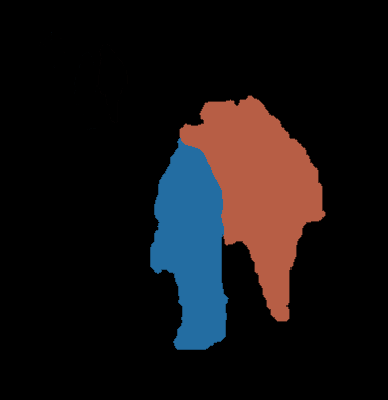}
\includegraphics[width=.09\textwidth]{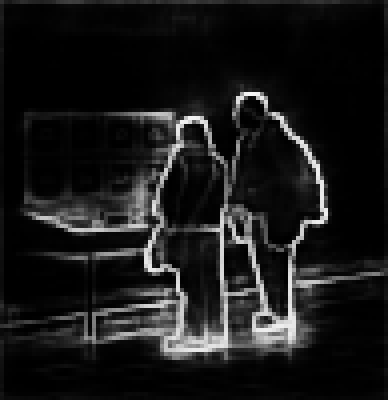}
\includegraphics[width=.09\textwidth]{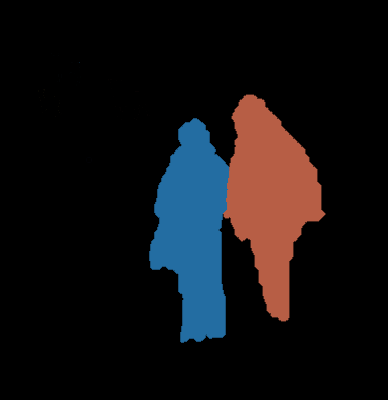}
\centering
\includegraphics[width=.09\textwidth,height=.092\textwidth]{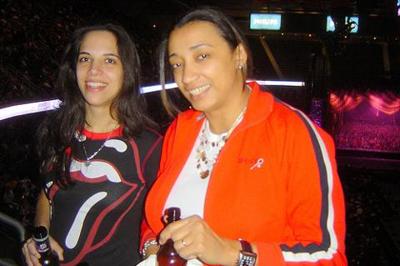}
\includegraphics[width=.09\textwidth,height=.092\textwidth]{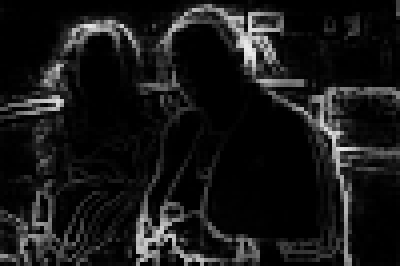}
\includegraphics[width=.09\textwidth,height=.092\textwidth]{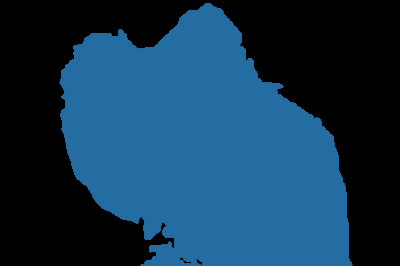}
\includegraphics[width=.09\textwidth,height=.092\textwidth]{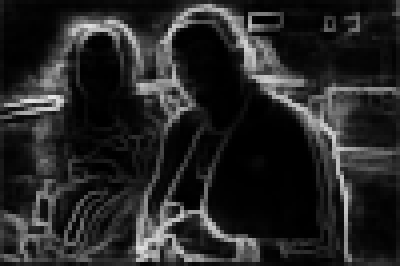}
\includegraphics[width=.09\textwidth,height=.092\textwidth]{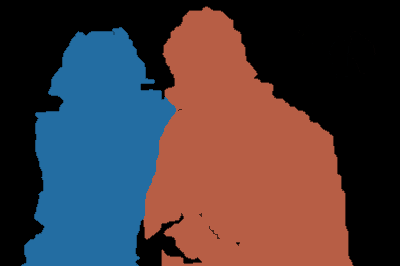}
\caption{Visualization of the boundary detection branch with and without the BE module.}
\label{fig:wbe}
\vspace{-2mm}
\end{figure}

Figure~\ref{fig:wbe} visualizes two examples of boundary detection and the corresponding salient instance detection with and without the BE module. We can see that our BE module helps detect the boundaries between objects, which is crucial to salient instance segmentation.

\subsection{Double Attention (DA) Module}\label{sec:da}
Detecting instance centroids and boundaries are two highly coupled sub-tasks, \ie, they can influence each other and further affect the SID performance. To efficiently learn these two sub-tasks, we propose the Double Attention (DA) module. Its design is based on two observations. First, since salient instances may have various shapes, we thus need to capture long-range spatial contextual information. Second, cross-class ambiguities of pseudo affinity labels influence both sub-tasks, while the class information from the channel-wise contexts can help address this problem. Hence, we combine channel-wise and spatial-wise attention mechanisms and organize them in parallel to form our DA module. We apply the DA module to both the centroid detection and the boundary detection branches, and share their weights. Unlike existing dual attention mechanisms~\cite{woo2018cbam,fu2019dual} that are only used to enhance the feature discriminatively, our DA module also allows information exchanges across these two branches, resulting in an improvement on both sub-tasks.

\begin{figure}[h]
\begin{center}
\includegraphics[width=0.6\linewidth]{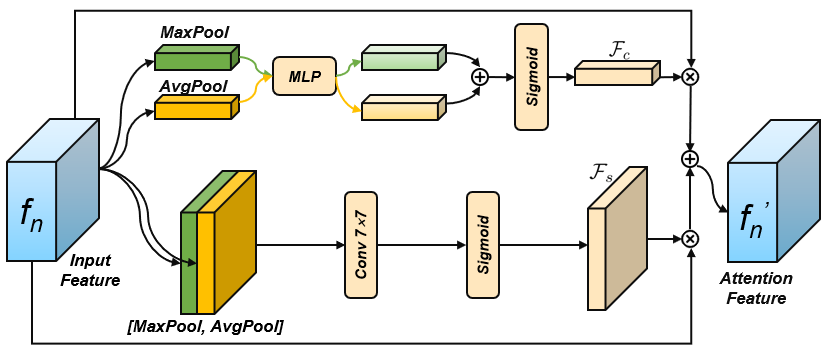}
\end{center}
\vspace{-3mm}
\centering
\caption{\tx{Double Attention module.}}
\label{fig:MA}
\vspace{-3mm}
\end{figure}

\tx{
Figure~\ref{fig:MA} shows the structure of our DA module. The top and bottom branches are channel-wise and spatial-wise attention blocks, respectively. Specifically, given the input features ${f}_{n}$, we compute the channel-wise attention features $\mathcal{F}_{c}$ as: $\mathcal{F}_{c} = \sigma(MLP(AvgPool_{c}({{f}_{n}}))$ \\$+MLP(MaxPool_{c}({{f}_{n}})))$,
where $MaxPool_{c}$ and $AvgPool_{c}$ denote two channel-wise pooling operations, and MLP is the multi-layer perception with one hidden layer to generate the attention features. We also compute the spatial-wise attention features $\mathcal{F}_{s}$ as:
$\mathcal{F}_{s} = \sigma(Conv_{7\times7}$ \\$([AvgPool_{s}({{f}_{n}}); MaxPool_{s}({{f}_{n}})]))$,
where $Conv_{7\times7}$ is a convolutional layer with kernel size 7. The final attention features ${{f}_{n}}^{'}$ are then computed as: ${{f}_{n}}^{'} = {{f}_{n}}\times\mathcal{F}_{c} + {{f}_{n}}\times\mathcal{F}_{s}$,
where $\times$ denotes the dot product operation, and $+$ is the element-wise summation operation.
}


\begin{figure}[h]
\centering
\begin{minipage}[t]{0.1\textwidth}
\centering
\scriptsize{\textbf{input\\ image}}
\end{minipage}
\begin{minipage}[t]{0.1\textwidth}
\centering
\scriptsize{\textbf{boundary \\(w/o DA)}}
\end{minipage}
\begin{minipage}[t]{0.1\textwidth}
\centering
\scriptsize{\textbf{centroid \\ (w/o DA)}}
\end{minipage}
\begin{minipage}[t]{0.1\textwidth}
\centering
\scriptsize{\textbf{instance mask \\(w/o DA)}}
\end{minipage}
\centering
\begin{minipage}[t]{0.1\textwidth}
\centering
\scriptsize{\textbf{ground\\ truth}}
\end{minipage}
\begin{minipage}[t]{0.1\textwidth}
\centering
\scriptsize{\textbf{boundary \\(w/ DA)}}
\end{minipage}
\begin{minipage}[t]{0.1\textwidth}
\centering
\scriptsize{\textbf{centroid \\ (w/ DA)}}
\end{minipage}
\begin{minipage}[t]{0.1\textwidth}
\centering
\scriptsize{\textbf{instance mask \\(w/ DA)}}
\end{minipage}
\\
\centering
\includegraphics[width=.1\textwidth]{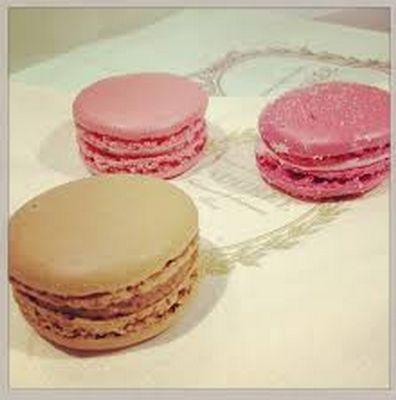}
\includegraphics[width=.1\textwidth]{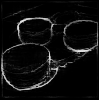}
\includegraphics[width=.1\textwidth]{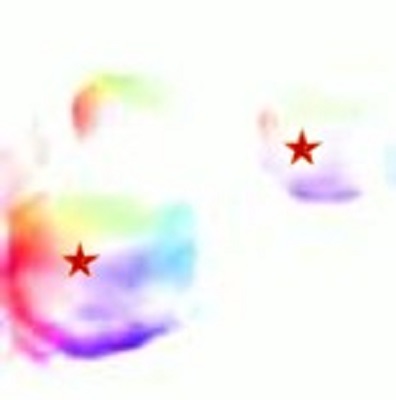}
\includegraphics[width=.1\textwidth]{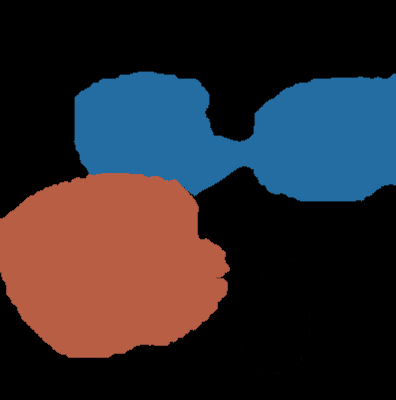}
\centering
\includegraphics[width=.1\textwidth]{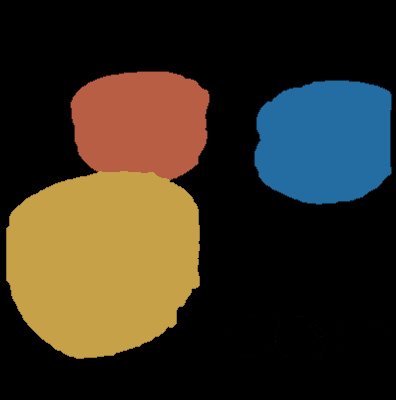}
\includegraphics[width=.1\textwidth]{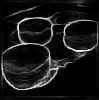}
\includegraphics[width=.1\textwidth]{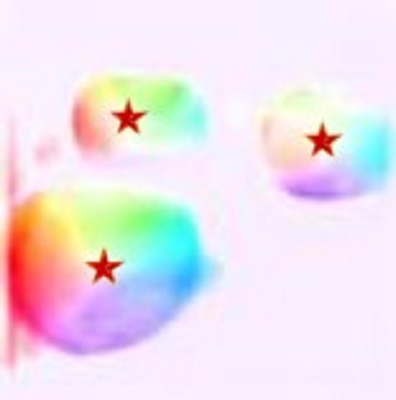}
\includegraphics[width=.1\textwidth]{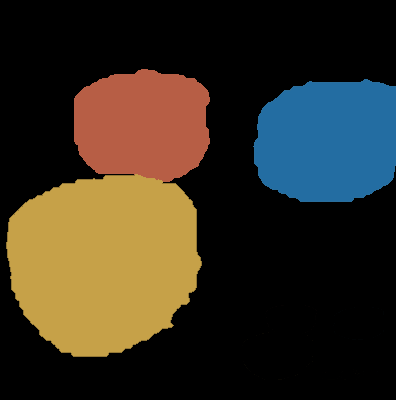}
\vspace{1mm}
\caption{Visualization of the DA module with and without the DA module.}\label{fig:da}
\vspace{-5mm}
\end{figure}

Figure~\ref{fig:da} shows the effectiveness of the proposed Double Attention module in enhancing the boundary and centroid detection performances.


%
%


\section{Experiments}
\vspace{-2mm}
\subsection{{Training and evaluation details}}
{\bf \tx{Datasets and metric}.} Our full network is trained on two image-level labels, class and subitizing. We use PASCAL VOC 2012 \cite{Everingham2010-pascal-voc-IJCV}, which is a semantic and instance segmentation dataset. However, we only use its class labels to train our network. The training set contains a total of 10,582 images of 20 classes. ILSO \cite{li2017instance} is a SID dataset. It contains 500 images with instance-aware pixel labels for training. For our weakly-supervised training, we extract the numbers of salient instances of these images as our subitizing labels. We perform evaluations on the test set of ILSO~\cite{li2017instance}, which has 300 images with ground truth instance masks. We use the mean Average Precision (mAP) metric~\cite{hariharan2014simultaneous} to evaluate the SID performance.
\\
\tx{\bf Training and Inference.} \kk{We train the proposed network separately. We train the centroid detection branch using the proposed centroid-based subitizing loss together with the centroid loss introduced in~\cite{jiwoon2019weakly,Neven2019InstanceSB}.
We train the boundary detection branch using the boundary loss introduced in~\cite{Ahn_2018_CVPR,jiwoon2019weakly}.
To train the saliency detection branch, we follow existing weakly-supervised SOD methods to use pseudo masks derived from class labels. Specifically, we first compute class activation maps via~\cite{zhou2016learning}. We then feed these maps together with the input image to a Conditional Random Field~\cite{krahenbuhl2011efficient} to generate pseudo object maps, and use these pixel-level pseudo labels to train the saliency detection branch.}
\kk{During inference, given an input image, WSID-Net first computes the centroids, boundaries, and saliency maps. To segment each salient instance, a random walk algorithm is used to detect salient regions, starting from the detected centroids until reaching the boundaries.}
\vspace{-4mm}

\subsection{{Implementation details}}
\tx{We implement WSID-Net on the Pytorch framework~\cite{paszke2019pytorch}.
Both training and testing are performed on a PC with an i7 4GHz CPU and a GTX 1080Ti GPU.
CRF is used to generate or refine pseudo labels. The hyper parameters of CRF are set as $w_{1}=4.0$, $w_{2}=3.0$, $\sigma_{\alpha}=49.0$, $\sigma_{\beta}=5.0$, and $\sigma_{\gamma}=3.0$.
We choose ResNet50 as the backbone for all three branches in WSID-Net. The backbone is initialized as in \cite{simonyan2014very}.
Input images are resized to 512$\times$512 resolution.
To minimize the loss function, we use the SGD optimizer with batch size 6 and initial learning rate 0.01. The learning rate decreases following poly policy ($lr_{itr} = lr_{init}(1-\frac{itr}{max_itr})^{\gamma}$). We train our WSID-Net for 5 epoches.}

\vspace{-2mm}
\subsection{{Comparion with the State-of-the-art Methods}}\label{sec42}
As we are the first to propose a weakly-supervised SID method, \kk{we compare our method to 2 existing fully-supervised state-of-the-art SID methods: S4Net~\cite{fan2019s4net} and MSRNet~\cite{li2017instance}.
We also prepare the following baselines from related tasks for evaluation. We choose 6 state-of-the-art weakly-supervised methods, with two from the SOD task C2SNet \cite{li2018contour} and NLDF~\cite{luo2017non}; one from the SID task MAP~\cite{zhang2016unconstrained}; one from the object detection (OD) task, DeepMask \cite{pinheiro2015learning}; and two from the Semantic Instance Segmentation task, PRM+D~\cite{cholakkal2019object} and IRN \cite{jiwoon2019weakly}. We adapt them by adding different post-processing strategies to these methods for deriving instance-level saliency maps from their original outputs, or modifying their networks and retrain them using our training data. Details are summarized as follows:}
\begin{itemize}
\item \kk{C2SNet~\cite{li2018contour} and NLDF~\cite{luo2017non} are proposed for salient object detection with contour prediction. We apply the MCG method~\cite{APBMM2014}, which takes a contour map as input and outputs segment proposals, to obtain multiple salient instance proposals, and then use MAP~\cite{zhang2016unconstrained} to filter out proposals with low confidences.}
\item MAP~\cite{zhang2016unconstrained} is a fully-supervised SID method, which learns to predict the bounding boxes of salient instances. Since it cannot output salient instance masks, we feed both the image and the bounding box to a CRF~\cite{krahenbuhl2011efficient} to obtain the segmented masks.

\item DeepMask~\cite{pinheiro2015learning} learns to predict class-agnostic segment proposals with object scores. We utilize a weakly-supervised SOD method WSS~\cite{wang2017learning} to filter out non-salient segment proposals by calculating the IoU between the object mask and the salient mask, and set the IoU threshold to 0.75.

\item IRN~\cite{jiwoon2019weakly} learns to predict class-specific segment proposals. We utilize the same filtering method as in DeepMask to select salient instances.

\item PRM+D~\cite{cholakkal2019object} is trained with class and per-class subitizing labels. \kk{We add one additional convolutional layer at the end of the network to merge their per-class outputs (originally 20 output maps for 20 classes) into one class-agnostic map, and then retrain it using our training data.}
\end{itemize}

\subsection{Performance Evaluation}
\begin{figure*}[!t]
\centering
\begin{minipage}[t]{0.090\textwidth}
\centering
\tiny{\textbf{Image}}
\end{minipage}
\begin{minipage}[t]{0.090\textwidth}
\centering
\tiny{\textbf{PRM+D \cite{cholakkal2019object}}}
\end{minipage}
\begin{minipage}[t]{0.090\textwidth}
\centering
\tiny{\textbf{DeepMask\cite{pinheiro2015learning}}}
\end{minipage}
\begin{minipage}[t]{0.090\textwidth}
\centering
\tiny{\textbf{C2SNet \cite{li2018contour}}}
\end{minipage}
\begin{minipage}[t]{0.090\textwidth}
\centering
\tiny{\textbf{IRN \cite{jiwoon2019weakly}}}
\end{minipage}
\begin{minipage}[t]{0.090\textwidth}
\centering
\tiny{\textbf{NLDF \cite{luo2017non}}}
\end{minipage}
\begin{minipage}[t]{0.090\textwidth}
\centering
\tiny{\textbf{MAP \cite{zhang2016unconstrained}}}
\end{minipage}
\begin{minipage}[t]{0.090\textwidth}
\centering
\tiny{\textbf{S4Net \cite{fan2019s4net}}}
\end{minipage}
\begin{minipage}[t]{0.090\textwidth}
\centering
\tiny{\textbf{WSID-Net (Ours)}}
\end{minipage}
\centering
\begin{minipage}[t]{0.090\textwidth}
\centering
\tiny{\textbf{GT}}
\end{minipage}
\\
\centering
\includegraphics[width=.090\textwidth]{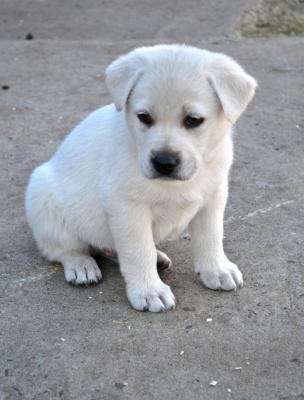}
\includegraphics[width=.090\textwidth]{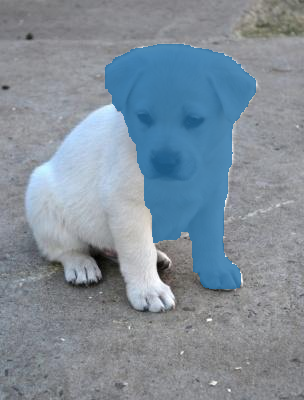}
\includegraphics[width=.090\textwidth]{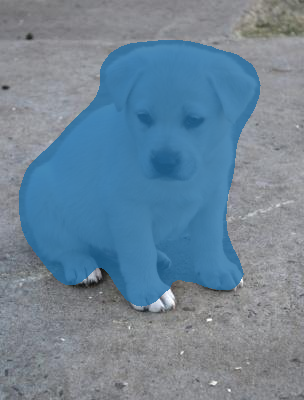}
\includegraphics[width=.090\textwidth]{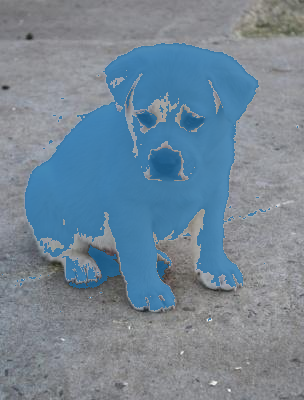}
\includegraphics[width=.090\textwidth]{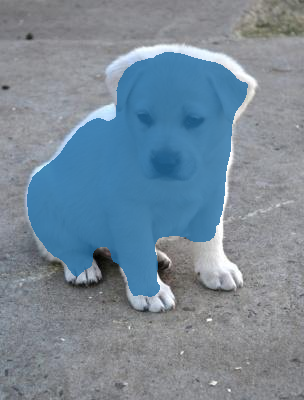}
\includegraphics[width=.090\textwidth]{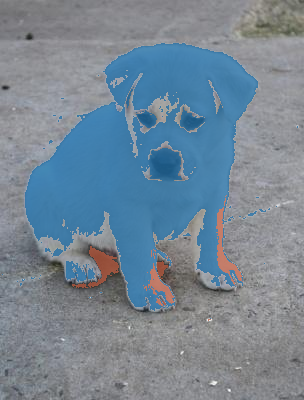}
\includegraphics[width=.090\textwidth]{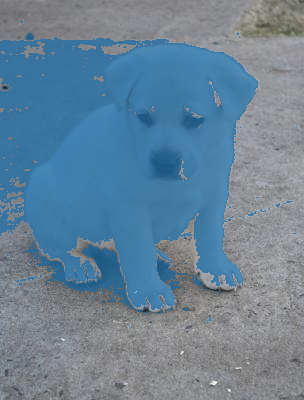}
\includegraphics[width=.090\textwidth]{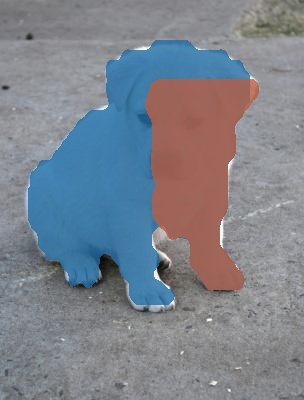}
\includegraphics[width=.090\textwidth]{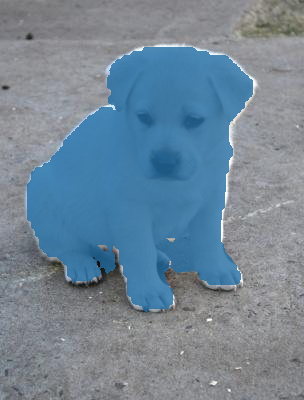}
\includegraphics[width=.090\textwidth]{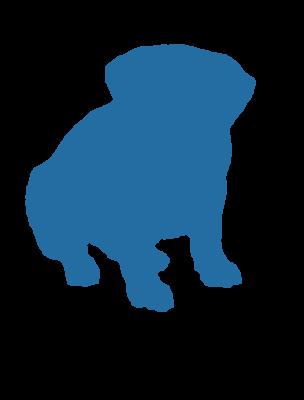}
\centering
\\
\includegraphics[width=.090\textwidth]{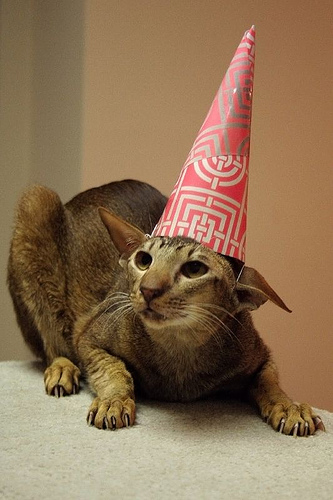}
\includegraphics[width=.090\textwidth]{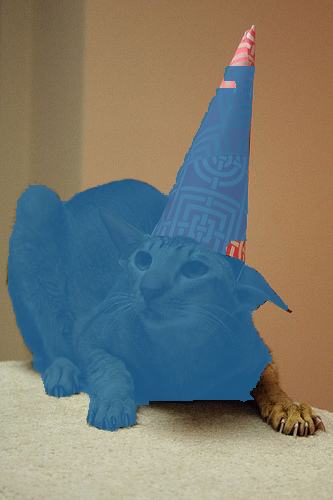}
\includegraphics[width=.090\textwidth]{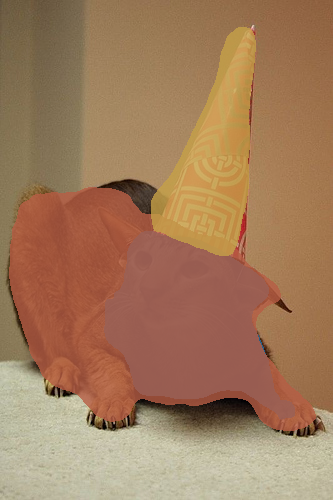}
\includegraphics[width=.090\textwidth]{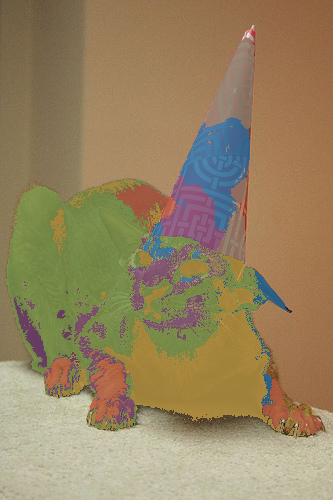}
\includegraphics[width=.090\textwidth]{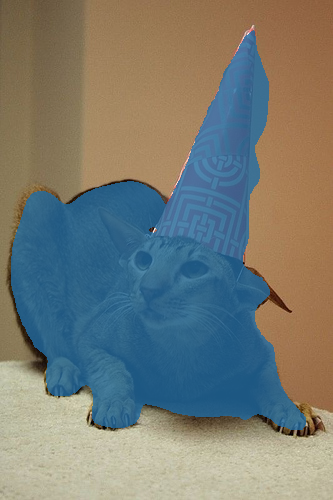}
\includegraphics[width=.090\textwidth]{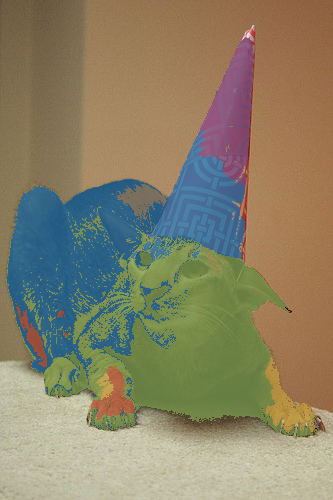}
\includegraphics[width=.090\textwidth]{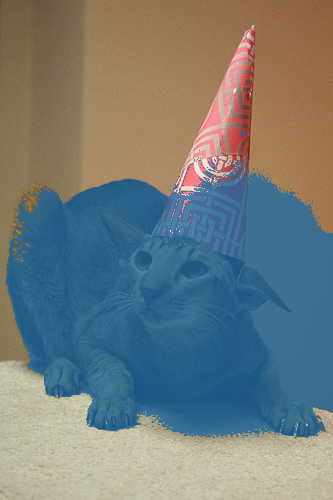}
\includegraphics[width=.090\textwidth]{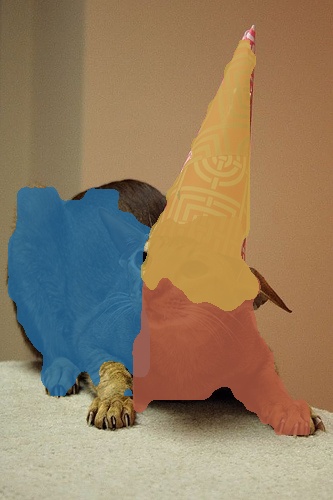}
\includegraphics[width=.090\textwidth]{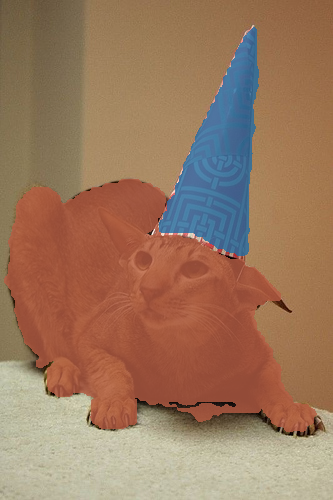}
\includegraphics[width=.090\textwidth]{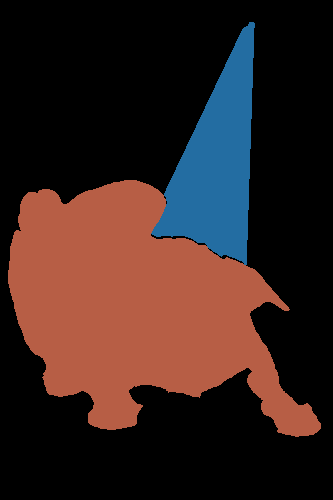}
\centering
\\
\includegraphics[width=.090\textwidth]{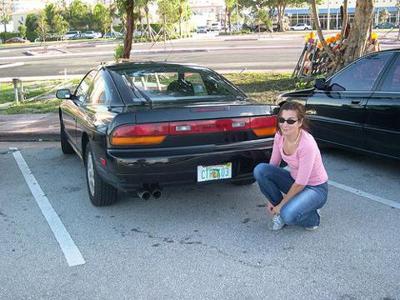}
\includegraphics[width=.090\textwidth]{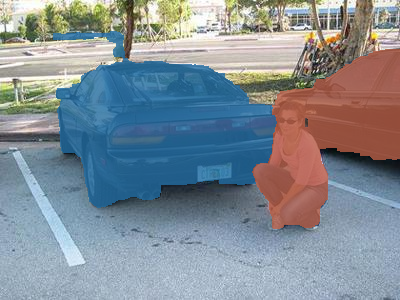}
\includegraphics[width=.090\textwidth]{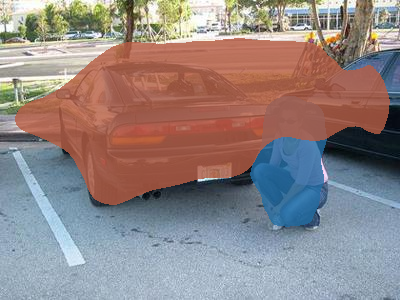}
\includegraphics[width=.090\textwidth]{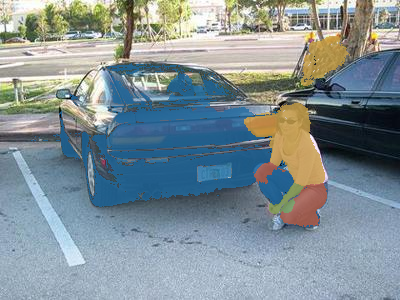}
\includegraphics[width=.090\textwidth]{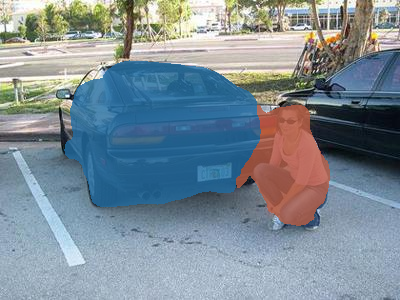}
\includegraphics[width=.090\textwidth]{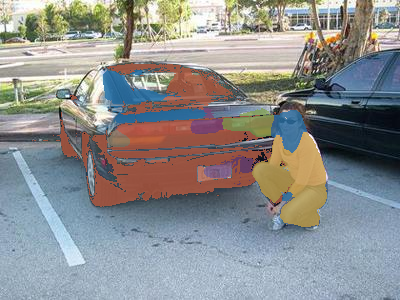}
\includegraphics[width=.090\textwidth]{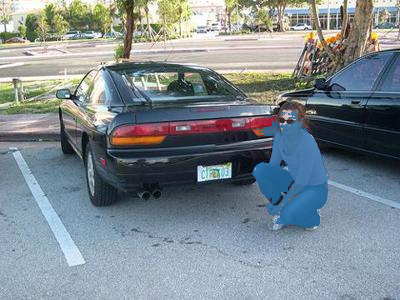}
\includegraphics[width=.090\textwidth]{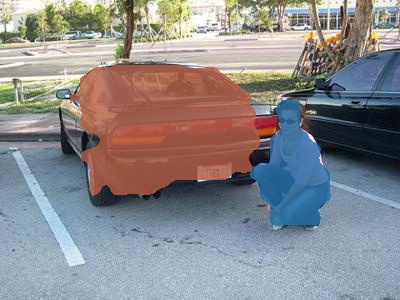}
\includegraphics[width=.090\textwidth]{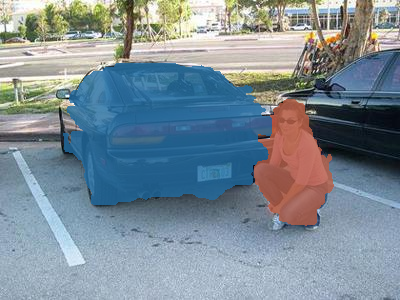}
\includegraphics[width=.090\textwidth]{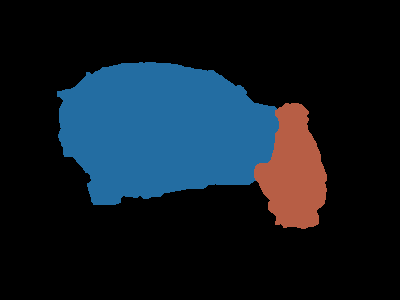}
\centering
\\
\includegraphics[width=.090\textwidth]{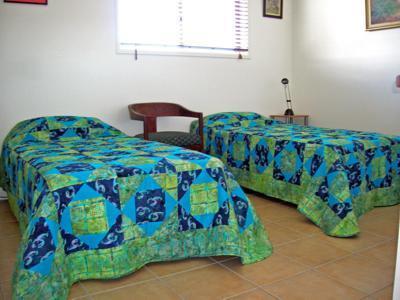}
\includegraphics[width=.090\textwidth]{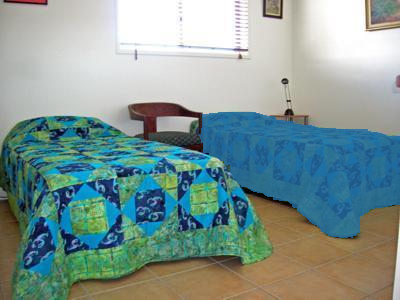}
\includegraphics[width=.090\textwidth]{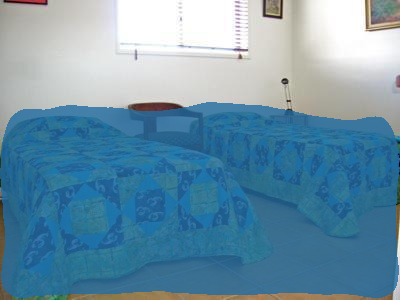}
\includegraphics[width=.090\textwidth]{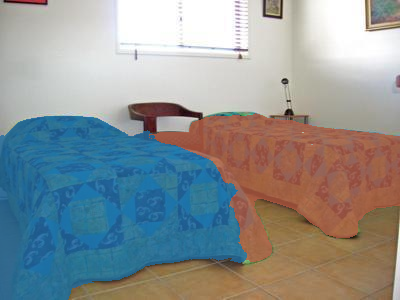}
\includegraphics[width=.090\textwidth]{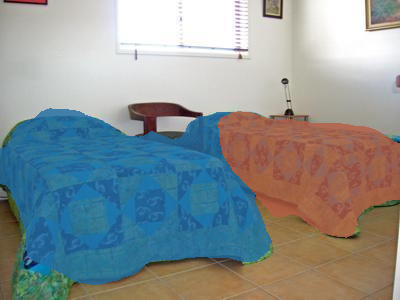}
\includegraphics[width=.090\textwidth]{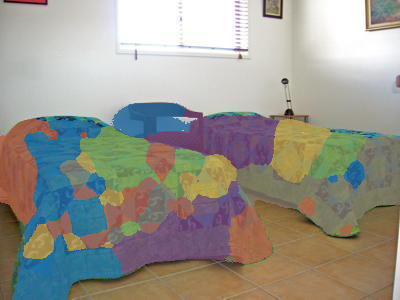}
\includegraphics[width=.090\textwidth]{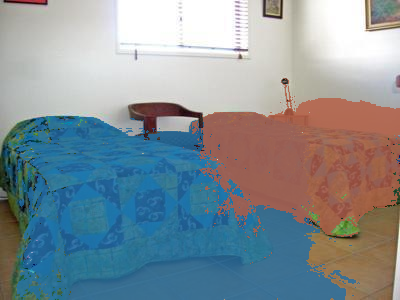}
\includegraphics[width=.090\textwidth]{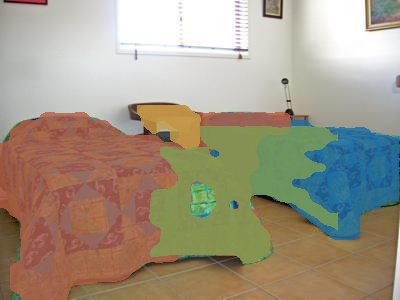}
\includegraphics[width=.090\textwidth]{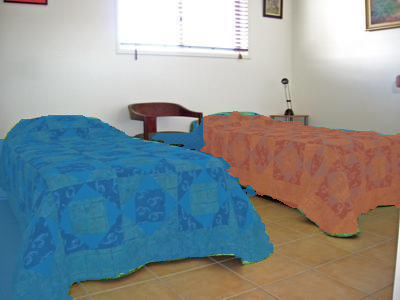}
\includegraphics[width=.090\textwidth]{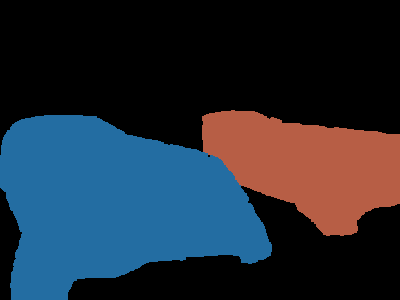}
\\
\centering
\includegraphics[width=.090\textwidth]{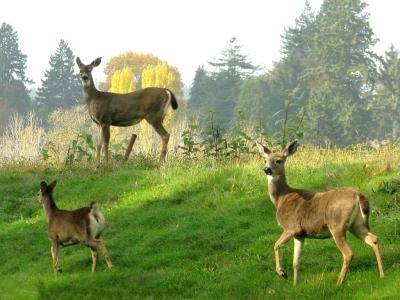}
\includegraphics[width=.090\textwidth]{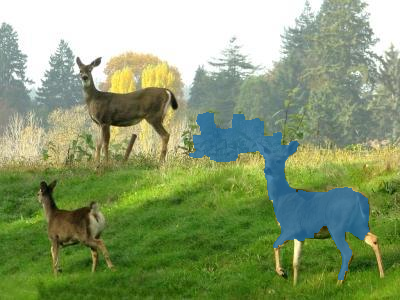}
\includegraphics[width=.090\textwidth]{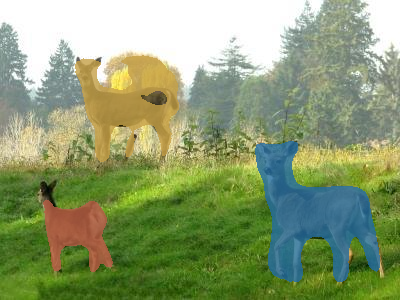}
\includegraphics[width=.090\textwidth]{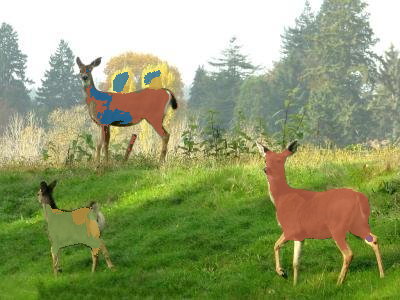}
\includegraphics[width=.090\textwidth]{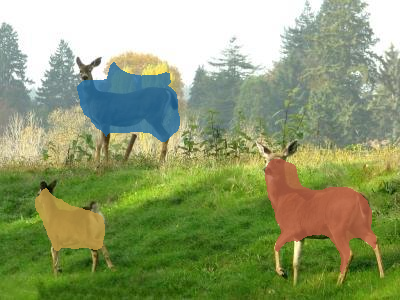}
\includegraphics[width=.090\textwidth]{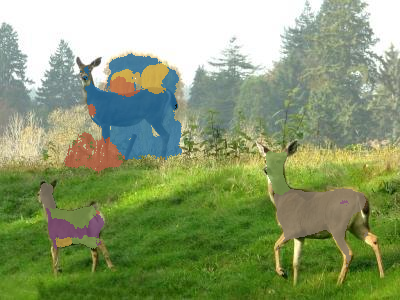}
\includegraphics[width=.090\textwidth]{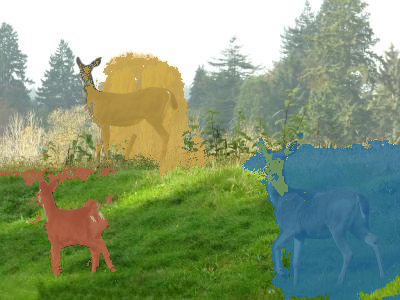}
\includegraphics[width=.090\textwidth]{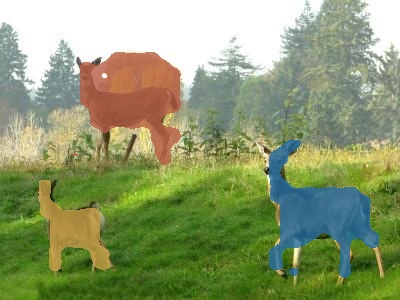}
\includegraphics[width=.090\textwidth]{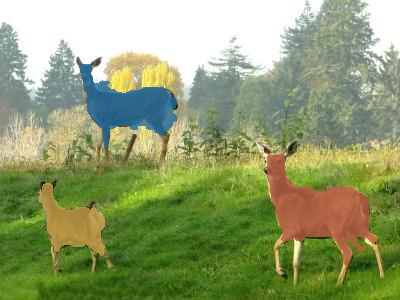}
\includegraphics[width=.090\textwidth]{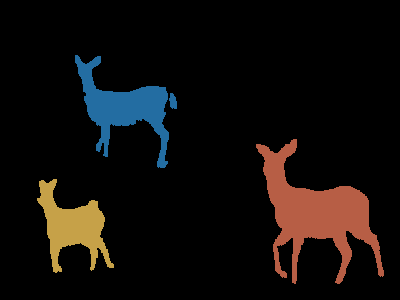}
\\
\centering
\includegraphics[width=.090\textwidth]{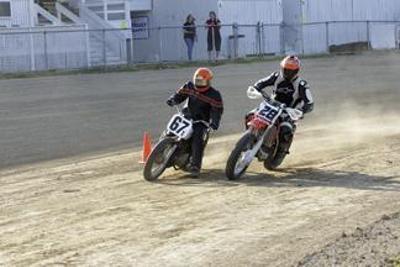}
\includegraphics[width=.090\textwidth]{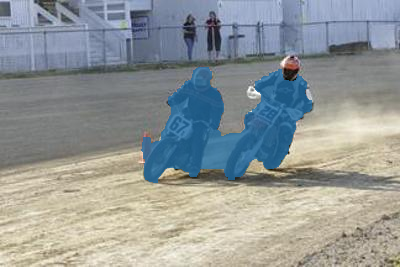}
\includegraphics[width=.090\textwidth]{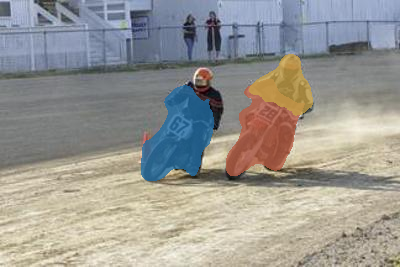}
\includegraphics[width=.090\textwidth]{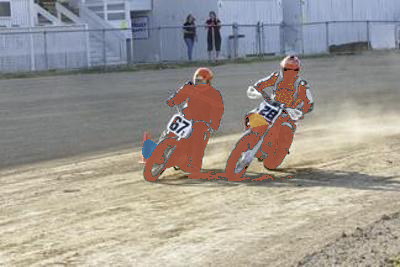}
\includegraphics[width=.090\textwidth]{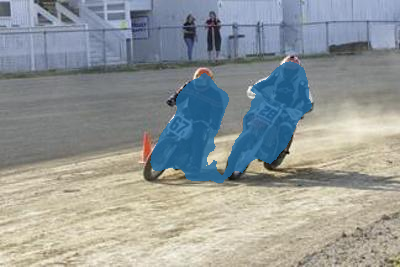}
\includegraphics[width=.090\textwidth]{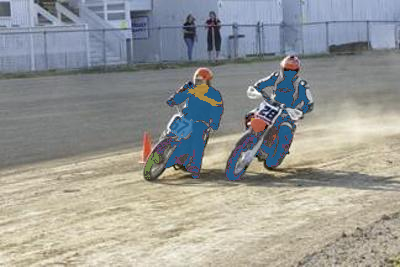}
\includegraphics[width=.090\textwidth]{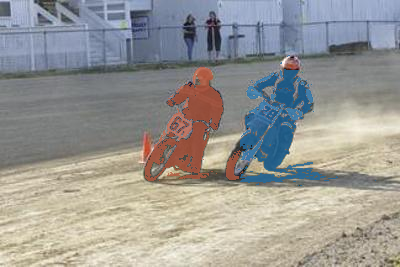}
\includegraphics[width=.090\textwidth]{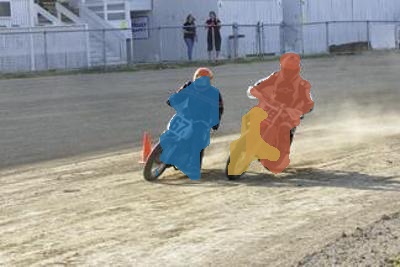}
\includegraphics[width=.090\textwidth]{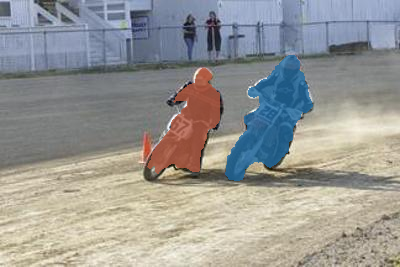}
\includegraphics[width=.090\textwidth]{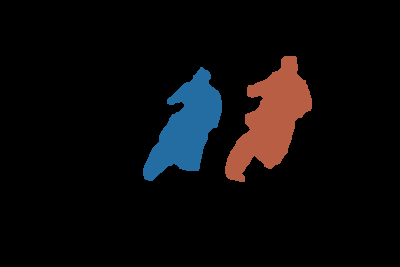}
\\
\centering
\includegraphics[width=.090\textwidth]{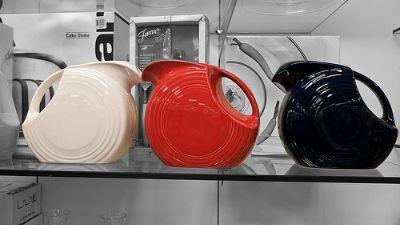}
\includegraphics[width=.090\textwidth]{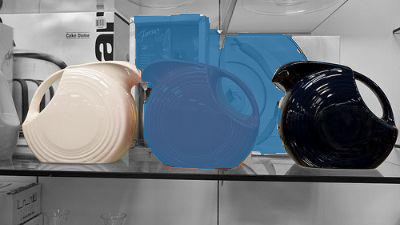}
\includegraphics[width=.090\textwidth]{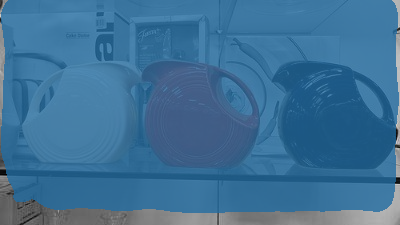}
\includegraphics[width=.090\textwidth]{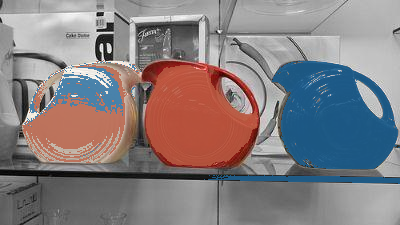}
\includegraphics[width=.090\textwidth]{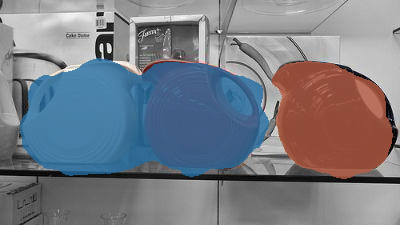}
\includegraphics[width=.090\textwidth]{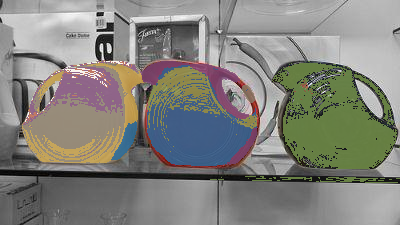}
\includegraphics[width=.090\textwidth]{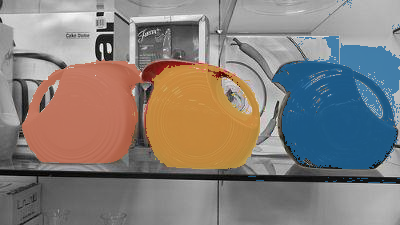}
\includegraphics[width=.090\textwidth]{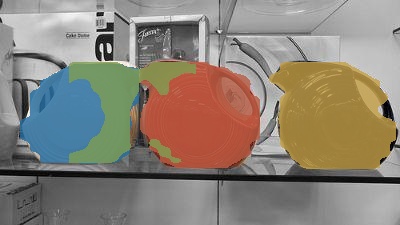}
\includegraphics[width=.090\textwidth]{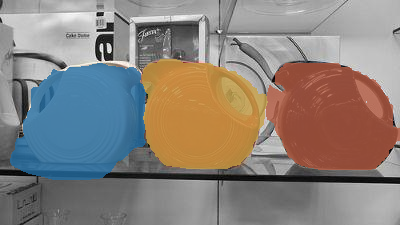}
\includegraphics[width=.090\textwidth]{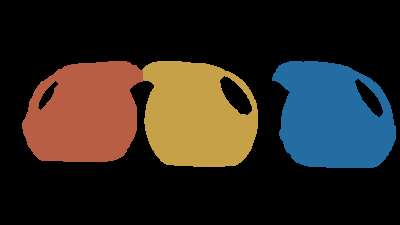}
\\
\centering
\includegraphics[width=.090\textwidth]{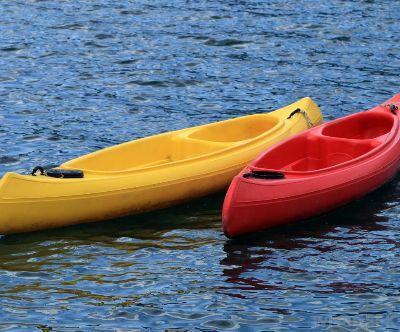}
\includegraphics[width=.090\textwidth]{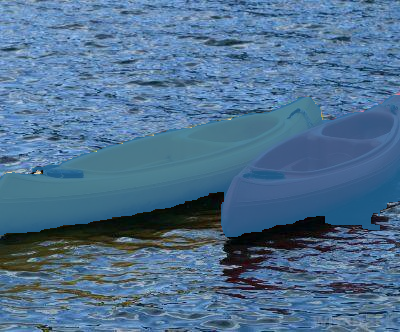}
\includegraphics[width=.090\textwidth]{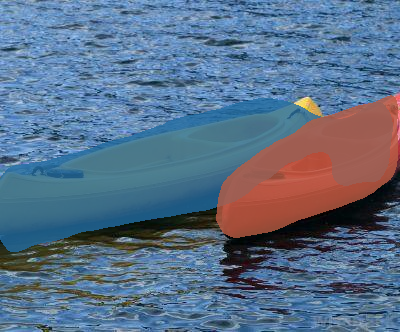}
\includegraphics[width=.090\textwidth]{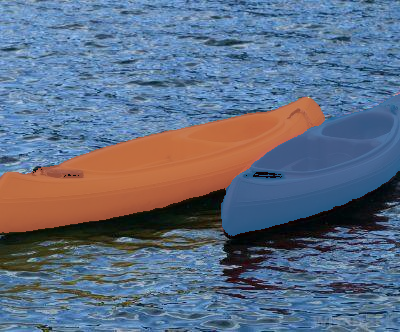}
\includegraphics[width=.090\textwidth]{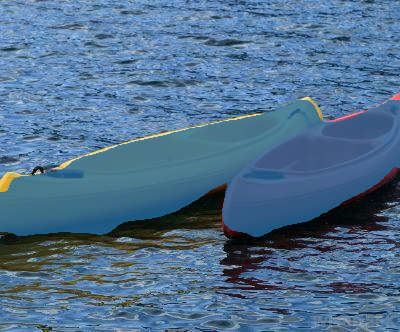}
\includegraphics[width=.090\textwidth]{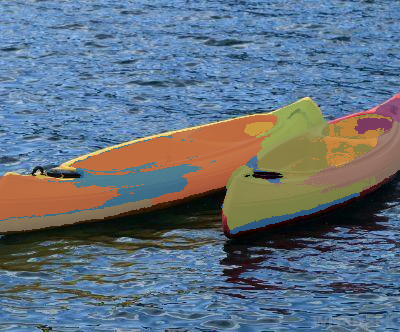}
\includegraphics[width=.090\textwidth]{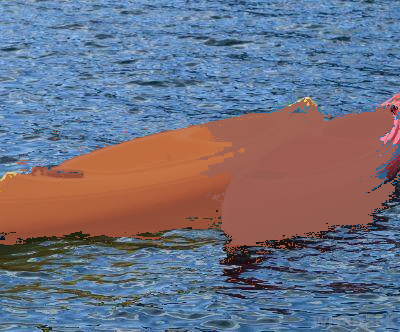}
\includegraphics[width=.090\textwidth]{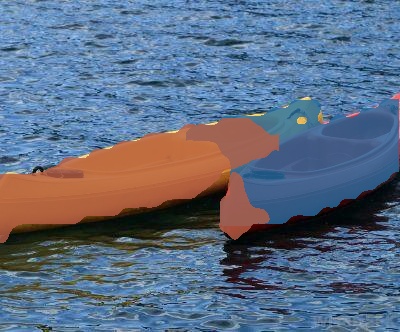}
\includegraphics[width=.090\textwidth]{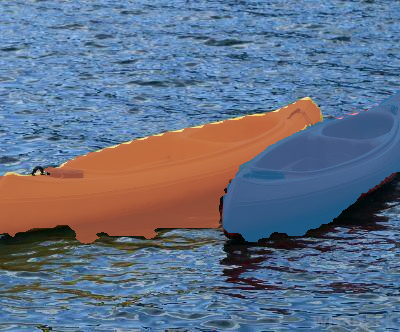}
\includegraphics[width=.090\textwidth]{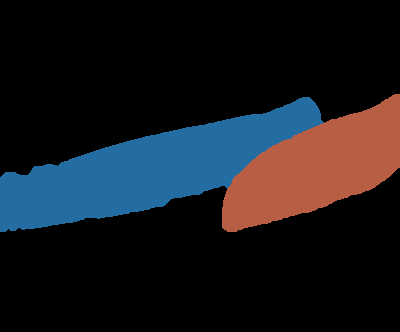}
\centering
\\
\includegraphics[width=.090\textwidth]{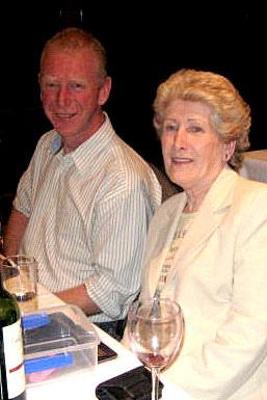}
\includegraphics[width=.090\textwidth]{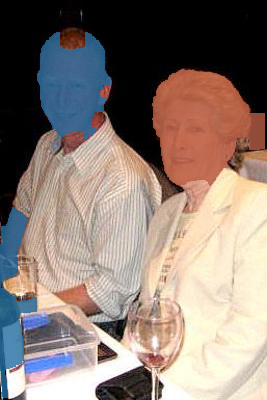}
\includegraphics[width=.090\textwidth]{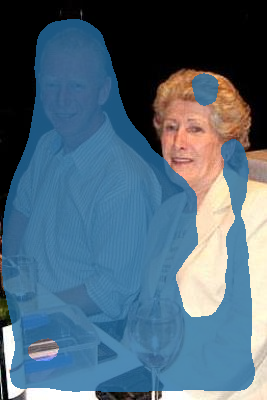}
\includegraphics[width=.090\textwidth]{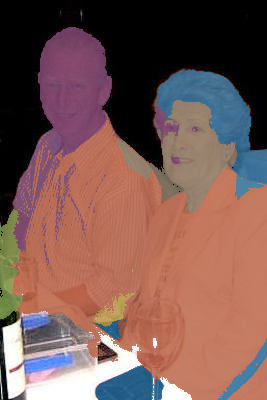}
\includegraphics[width=.090\textwidth]{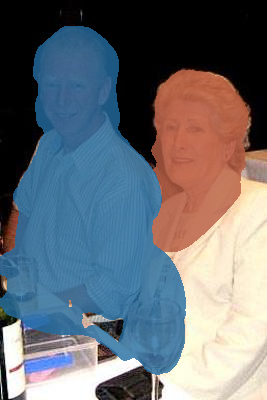}
\includegraphics[width=.090\textwidth]{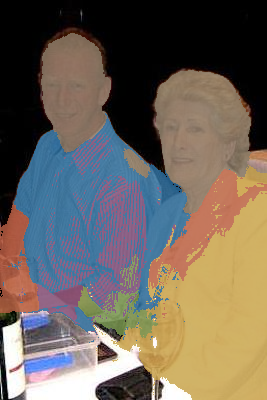}
\includegraphics[width=.090\textwidth]{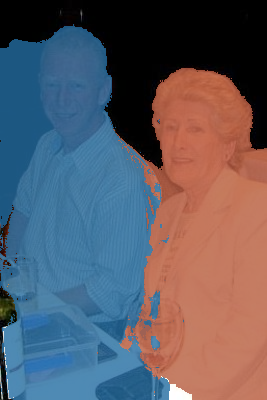}
\includegraphics[width=.090\textwidth]{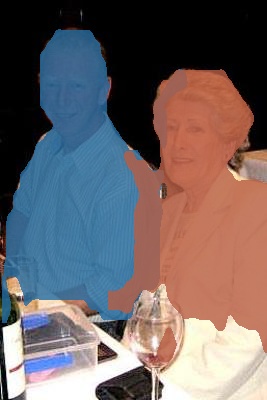}
\includegraphics[width=.090\textwidth]{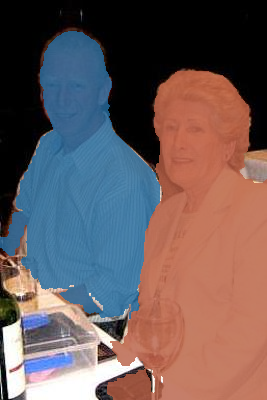}
\includegraphics[width=.090\textwidth]{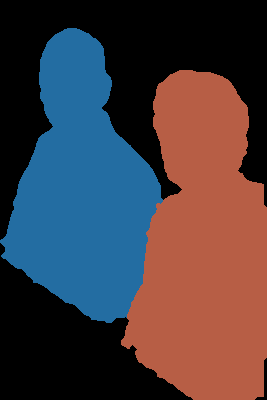}
\caption{Qualitative results of our method, compared with existing fully-supervised methods (S4Net\cite{fan2019s4net} and MAP~\cite{zhang2016unconstrained}) and modified baselines (PRM+D~\cite{cholakkal2019object}, DeepMask~\cite{pinheiro2015learning}, C2SNet~\cite{li2018contour}, NLDF~\cite{luo2017non}, and IRN~\cite{jiwoon2019weakly}). Refer to Section~\ref{sec42} and Table~\ref{tab:SID} on how we modify and train these baselines, in order to perform appropriate comparison. }
\label{fig:instance}
\vspace{-4mm}
\end{figure*}

\tx{\bf Quantitative evaluation.} We quantitatively evaluate our method in Table~\ref{tab:SID}\tx{$\dagger$}. mAP@0.7 is the most difficult metric as it requires the IoU value to be over 70\%. Our method achieves a better performance of about 10\% over the second-place weakly-supervised baseline. \tx{These results show that our method achieves the best performance using just two types of image-level labels}.
{\let\thefootnote\relax\footnote{{$^{\dagger}$ As of today, the codes for MSRNet~\cite{li2017instance} are still not available. Following~\cite{fan2019s4net}, we directly copy the numbers reported in~\cite{li2017instance} to our submission for a quantitative comparison.}}}
\\
\tx{\bf Qualitative evaluation.} We further qualitatively compare our method as shown in Figure~\ref{fig:instance}. Our method is able to delineate the instance boundaries clearly, and output an accurate number of segmented salient instances directly, as shown in column 9. In contrast, \tx{(1) }PRM+D and IRN fail to detect integral instances with inferior detected boundaries (\eg, rows 1 and 9); \tx{(2) }C2SNet and NLDF tend to recognize texture boundaries, resulting in fragmented instances (\eg, rows 2, 3 and 4); \tx{(3) }DeepMask and S4Net suffer from the over-detection problem, as they fail to distinguish instance proposals belonging to the same instance (\eg, row 2); \tx{(4) }and MAP is a bounding-box based method that fails to get clear instance boundary even post-processed by a widely adopted segmentation method,  CRF (\eg, rows 1 and 2). Overall, our method outperforms the baselines, as a result of the centroid-based subitizing loss and the carefully designed BE and DA modules.

\begin{table*}[h]
\begin{center}
 \begin{tabular}{||c c c c c c c||}
 \hline
 \textbf{\scriptsize{Methods}} & \textbf{\begin{tabular}[c]{@{}c@{}}\scriptsize{Original} \\ \scriptsize{task}\end{tabular}} & \textbf{\begin{tabular}[c]{@{}c@{}}\scriptsize{Supervision} \\ \scriptsize{types}\end{tabular}} & \textbf{\begin{tabular}[c]{@{}c@{}}\scriptsize{Training labels}\end{tabular}} & \textbf{\begin{tabular}[c]{@{}c@{}}\scriptsize{Auxiliary} \\\scriptsize{models}\end{tabular}} & \textbf{\begin{tabular}[c]{@{}c@{}}\scriptsize{mAP} \\ \scriptsize{@0.5$\uparrow$}\end{tabular}} & \textbf{\begin{tabular}[c]{@{}c@{}}\scriptsize{mAP} \\ \scriptsize{@0.7$\uparrow$}\end{tabular}} \\ [0.5ex]
 \hline\hline
 \scriptsize{MSRNet \cite{li2017instance}} & \scriptsize{SID} & \scriptsize{FS} & \begin{tabular}[c]{@{}c@{}}\scriptsize{object-level and} \\\scriptsize{instance-level pixel masks}\end{tabular} & \scriptsize{MAP \cite{zhang2016unconstrained}, MCG \cite{APBMM2014}} & \scriptsize{{65.3\%}} & \scriptsize{{52.3\%}} \\
 \hline
 \scriptsize{MAP \cite{zhang2016unconstrained}} & \scriptsize{SID} & \scriptsize{FS}  & \scriptsize{instance-level bounding box} & \scriptsize{N/A} & \scriptsize{56.6\%} & \scriptsize{24.8\%} \\
 \hline
 \scriptsize{S4Net \cite{fan2019s4net}} & \scriptsize{SID} & \scriptsize{FS} & \scriptsize{instance-level pixel mask} & \scriptsize{N/A} & \scriptsize{{82.2\%}} & \scriptsize{{59.6\%}} \\
 \hline
 \hline
 \scriptsize{C2SNet \cite{li2018contour}} & \scriptsize{SOD} & \scriptsize{WS} & \scriptsize{unlabeled images} & \begin{tabular}[c]{@{}c@{}}\scriptsize{CEDN \cite{Yang2016object},} \\ \scriptsize{MAP \cite{zhang2016unconstrained},} \scriptsize{MCG \cite{APBMM2014}}\end{tabular} & \scriptsize{41.1\%} & \scriptsize{25.4\%} \\
 \hline
 \scriptsize{NLDF \cite{luo2017non}} & \scriptsize{SOD} & \scriptsize{WS} & \begin{tabular}[c]{@{}c@{}}\scriptsize{object-level pixel mask}\end{tabular} & \scriptsize{MAP \cite{zhang2016unconstrained}, MCG \cite{APBMM2014}} & \scriptsize{45.5\%} & \scriptsize{24.5\%} \\
 \hline
 \scriptsize{DeepMask \cite{pinheiro2015learning}} & \scriptsize{OD} & \scriptsize{WS} & \scriptsize{instance-level bounding box} & \scriptsize{N/A} & \scriptsize{37.1\%} & \scriptsize{20.5\%} \\
 \hline
 \scriptsize{PRM+D \cite{cholakkal2019object}} & \scriptsize{SIS} & \scriptsize{WS} & \scriptsize{class, subitizing labels} & \scriptsize{MCG \cite{APBMM2014}} & \scriptsize{49.6\%} & \scriptsize{31.2\%} \\
 \hline
 \scriptsize{IRN \cite{jiwoon2019weakly}} & \scriptsize{SIS} & \scriptsize{WS} & \scriptsize{class label} & \scriptsize{N/A} & \scriptsize{57.1\%} & \scriptsize{37.4\%} \\
 \hline
 \scriptsize{Ours} & \scriptsize{SID} &  \scriptsize{WS}  & \scriptsize{class, subitizing labels} & \scriptsize{N/A} &   \textbf{\scriptsize{\textcolor{red}{61.9\%}}}  &  \textbf{\scriptsize{\textcolor{red}{47.2\%}}} \\
 \hline
\end{tabular}
\end{center}\vspace{-2mm}
\caption{Quantitative evaluation of our method against six baseline methods and state-of-the-art fully-supervised SID methods. For the compared methods, we show their original tasks, supervision types, training labels and auxiliary pre-trained models in the 2$nd$ to 5$th$ columns. SID, SOD, OD, SIS are short of salient instance detection, salient object detection, object detection and semantic instance segmentation, respectively. FS and WS denote Fully-Supervised and Weakly-Supervised. Best performances among the weakly-supervised methods are marked in \textbf{\textcolor{red}{red}}.
}%
\label{tab:SID}\vspace{-4mm}
\end{table*}

\vspace{-4mm3}
\subsection{Internal Analysis}
\tx{We first investigate} how our BE, DA modules, and $\mathcal{L}_{\mathcal{SU}}$ affect the SID performance. We provide ablation study on our model design based on the mAP@IoU metric. From the results in Table \ref{tab:ab1}, we can see that the SID performance is continuously increased as we incorporate these modules. This shows that these modules can help boost the performances of the centroid and boundary detection sub-tasks, which play a vital role in detecting salient instances. Figures~\ref{fig:wsu},~\ref{fig:wbe}, and~\ref{fig:da} provide additional visual comparisons to demonstrate the effectiveness of the BE, DA modules and the $\mathcal{L}_{\mathcal{SU}}$.

\tx{
We also evaluate the design choices of the DA module and the influence of using different backbones. Due to page limitation, we show these analytical results in the Supplemental.
}
\vspace{-5mm}
\section{Conclusion}
In this paper, we propose the first weakly-supervised SID method that is trained on class and subitizing labels. Our WSID-Net learns to predict object boundary, instance centroid, and salient region. By using the proposed Boundary Enhancement module, Double Attention module, and centroid-based subitizing loss, our method can identify and segment each salient instance effectively. Both quantitative and qualitative experiments demonstrate the effectiveness of the proposed method compared with baseline methods.

Our method does have limitations. It may fail when our saliency detection branch (as well as existing weakly-supervised SOD methods) cannot detect the majority of the salient regions, due to complex background textures and colors, as shown in Figure~\ref{fig:failure}. As a future work, we are exploring to incorporate a discriminative network of generative adversarial learning to improve the SOD performance of our saliency detection branch, and extend our method to handle videos.
\vspace{0.1in}

\begin{minipage}{0.45\textwidth}
\flushleft
\begin{tabular}{||c c c||}
\hline
\textbf{\scriptsize{method}} & \textbf{\scriptsize{mAP@0.5$\uparrow$}} & \textbf{\scriptsize{mAP@0.7$\uparrow$}} \\
\hline
\hline
\begin{tabular}[c]{@{}c@{}}\scriptsize{Ours} \\\scriptsize{(w/o DA, BE, $\mathcal{L}_{\mathcal{SU}}$)}\end{tabular} & \scriptsize{57.1\%} & \scriptsize{37.4\%} \\
\hline
\scriptsize{Ours (w/o DA)} & \scriptsize{60.3\%} & \scriptsize{45.1\%} \\
\hline
\scriptsize{Ours (w/o BE)} & \scriptsize{58.2\%} & \scriptsize{44.3\%} \\
\hline
\scriptsize{Ours (w/o $\mathcal{L}_{\mathcal{SU}}$)} & \scriptsize{59.9\%} & \scriptsize{45.0\%} \\
\hline
\scriptsize{Ours}  & \textbf{\scriptsize{61.9\%}} & \textbf{\scriptsize{47.2\%}} \\
\hline
\end{tabular}
\makeatletter\def\@captype{table}\makeatother\caption{Ablation study of WSID-Net.}
\label{tab:ab1}
\end{minipage}
\begin{minipage}{.55\textwidth}
\centering
\begin{minipage}[t]{0.2\textwidth}
\centering
\tiny{\textbf{image}}
\end{minipage}
\begin{minipage}[t]{0.2\textwidth}
\centering
\tiny{\textbf{centroid}}
\end{minipage}
\begin{minipage}[t]{0.2\textwidth}
\centering
\tiny{\textbf{our saliency}}
\end{minipage}
\begin{minipage}[t]{0.25\textwidth}
\centering
\tiny{\textbf{instance w/ our saliency}}
\end{minipage}
\centering
\includegraphics[width=.2\textwidth]{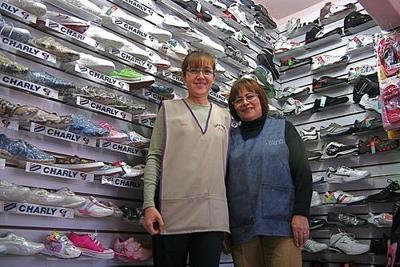}
\includegraphics[width=.2\textwidth]{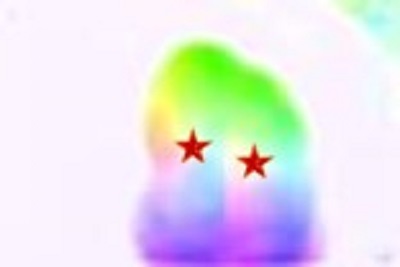}
\includegraphics[width=.2\textwidth]{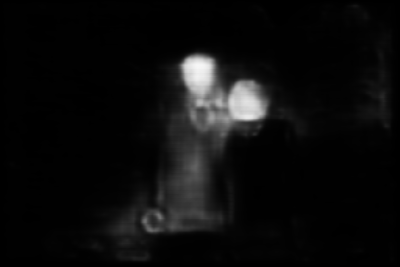}
\includegraphics[width=.2\textwidth]{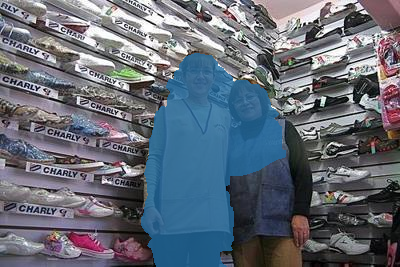}
\centering
\begin{minipage}[t]{0.2\textwidth}
\centering
\tiny{\textbf{GT of instance}}
\end{minipage}
\begin{minipage}[t]{0.2\textwidth}
\centering
\tiny{\textbf{boundary}}
\end{minipage}
\begin{minipage}[t]{0.2\textwidth}
\centering
\tiny{\textbf{GT of saliency}}
\end{minipage}
\begin{minipage}[t]{0.25\textwidth}
\centering
\tiny{\textbf{instance w/ GT saliency}}
\end{minipage}
\centering
\includegraphics[width=.2\textwidth]{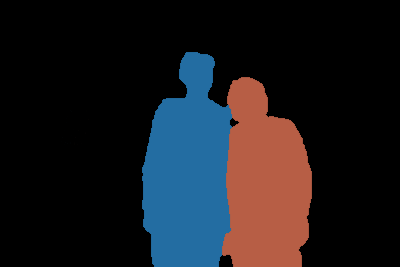}
\includegraphics[width=.2\textwidth]{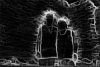}
\includegraphics[width=.2\textwidth]{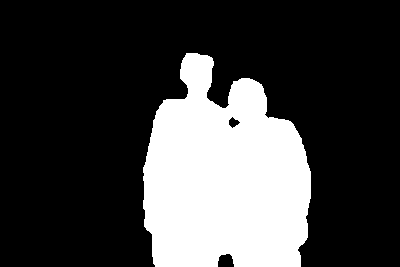}
\includegraphics[width=.2\textwidth]{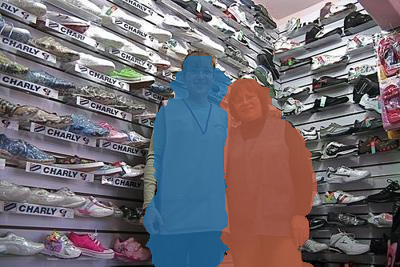}
\makeatletter\def\@captype{figure}\makeatother\caption{A failure case.}
\label{fig:failure}
\end{minipage}


\bibliography{egbib}
\end{document}